%% file: wast3d.tex
\documentclass[runningheads]{llncs}

\usepackage{eccv}

\usepackage{eccvabbrv}

\usepackage{graphicx}
\usepackage{booktabs}
\usepackage{placeins}
\usepackage{color,soul}
\usepackage{amsmath, bm}

\usepackage[accsupp]{axessibility}  

 \usepackage[pagebackref,breaklinks,colorlinks,citecolor=eccvblue]{hyperref}
\usepackage{hyperref}

\usepackage{orcidlink}

\begin{document}


\title{WaSt-3D: Wasserstein-2 Distance for Scene-to-Scene Stylization on 3D Gaussians} 

\titlerunning{WaSt-3D}

\author{
Dmytro Kotovenko\inst{1}\thanks{Work done during an internship at Meta.} \and
Olga Grebenkova\inst{1} \and
Nikolaos Sarafianos\inst{2} \and
Avinash Paliwal\inst{3} \and
Pingchuan Ma\inst{1} \and
Omid Poursaeed\inst{2} \and
Sreyas Mohan\inst{2} \and
Yuchen Fan\inst{2} \and
Yilei Li\inst{2} \and
Rakesh Ranjan\inst{2} \and
Björn Ommer\inst{1}
}

\authorrunning{D.~Kotovenko \textit{et al.}}

\institute{CompVis @ LMU Munich and MCML \and
Meta Reality Labs \qquad  \qquad 
\inst{3} Texas A\&M University
}

\maketitle

\begin{abstract}
  While style transfer techniques have been well-developed for 2D image stylization, the extension of these methods to 3D scenes remains relatively unexplored. Existing approaches demonstrate proficiency in transferring colors and textures but often struggle with replicating the geometry of the scenes. In our work, we leverage an explicit Gaussian Splatting (GS) representation and directly match the distributions of Gaussians between style and content scenes using the Earth Mover's Distance (EMD). By employing the entropy-regularized Wasserstein-2 distance, we ensure that the transformation maintains spatial smoothness. Additionally, we decompose the scene stylization problem into smaller chunks to enhance efficiency. This paradigm shift reframes stylization from a pure generative process driven by latent space losses to an explicit matching of distributions between two Gaussian representations. Our method achieves high-resolution 3D stylization by faithfully transferring details from 3D style scenes onto the content scene. Furthermore, WaSt-3D consistently delivers results across diverse content and style scenes without necessitating any training, as it relies solely on optimization-based techniques. See our project page for additional results and source code: \href{https://compvis.github.io/wast3d/}{https://compvis.github.io/wast3d/}.
\end{abstract}

\input{tex/introduction}

\input{tex/related_work}

\input{tex/Approach}

\input{tex/experiments}

\input{tex/conclusion}

\section*{Acknowledgements}
This project has been supported by the German Federal Ministry for Economic Affairs and Climate
Action within the project ``NXT GEN AI METHODS – Generative Methoden für Perzeption, Prädiktion und Planung'' and the German Research Foundation (DFG) project 421703927.  
The authors gratefully acknowledge the Gauss Center for Supercomputing for providing compute through the NIC on JUWELS at JSC and the HPC resources supplied by the Erlangen National High Performance Computing Center (NHR@FAU funded by DFG).
We also thank Ming Gui for the insightful discussion on the project and Owen Vincent for his support with technical questions.

\bibliographystyle{splncs04}
\bibliography{References}
\end{document}

%% file: tex/introduction.tex
\begin{figure}[t]
    \centering
     \includegraphics[width=1.\textwidth]{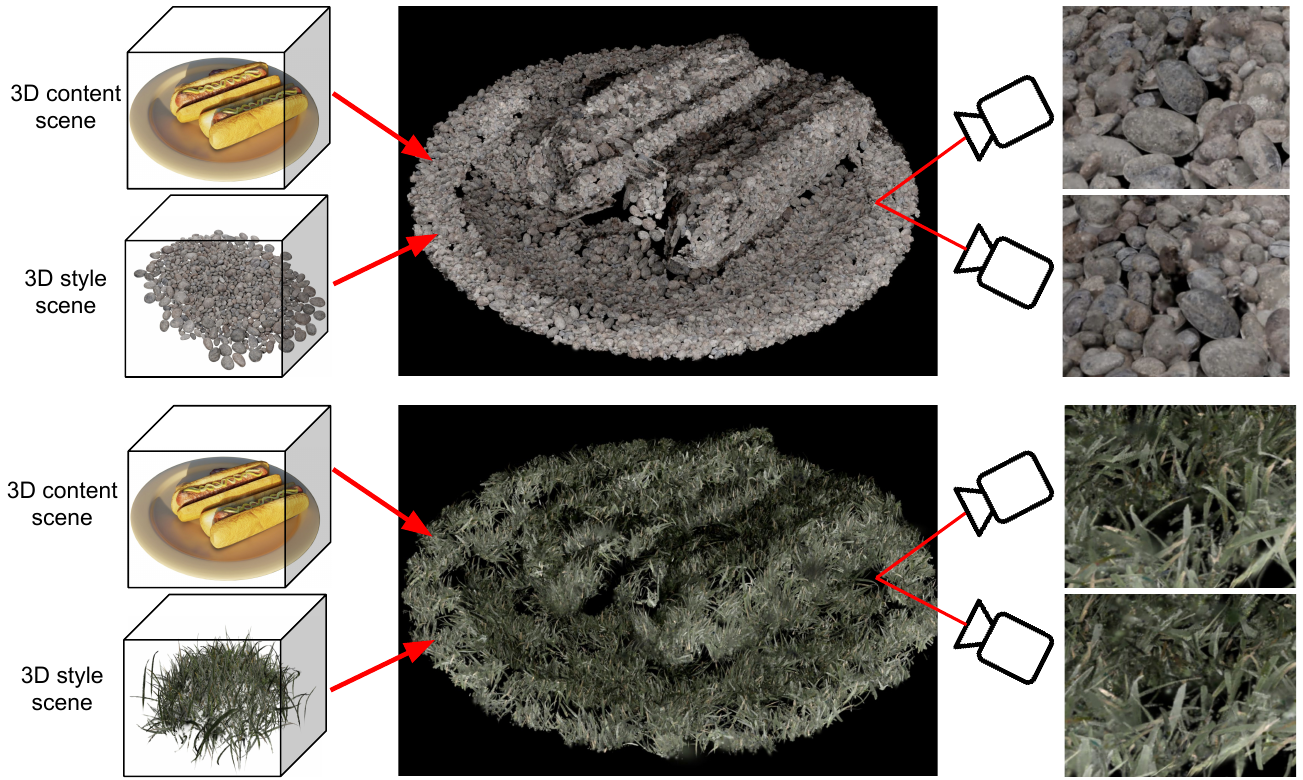}
    \caption{Our stylization of the hot dog scene using 2 different style scenes: \textit{pebbles} and \textit{grass}. Our method accepts as an input two pretrained Gaussian splattings scenes and merges them together by clustering content scene, selecting best style cluster and finally minimizing Sinkhorn divergence between the pairs. High resolution 3D details are depicted in the last column by rendering the same shape volume from two different viewpoints. }
    \label{fig:title}
\end{figure}

\section{Introduction}
Style transfer is a well researched method that allows to create artistic images, which combines content from one input source and style from another. This process has been widely used in 2D images to create visually appealing and unique results. However, as virtual reality and 3D approaches become more prevalent, there is a growing interest in adapting style transfer techniques to work in the three-dimensional space. 
Incorporating style transfer into 3D visuals opens up new possibilities for creating unique and engaging content. By merging the artistic styles of different sources with the three-dimensional aspects of virtual reality, creators can push the boundaries of what is possible in terms of visual storytelling and design. 

In computer graphics, stylized models serve as artistic interpretations of 3D objects, often featuring exaggerated or simplified elements reminiscent of cartoon characters, abstract shapes, or low-poly designs. This creative approach extends to 3D game design, where various art styles shape the visual identity of virtual worlds and gaming experiences. Realism and Fantasy Realism aim for lifelike authenticity, while Low Poly art embraces a minimalist aesthetic with simplified geometric shapes, and Cartoon art infuses games with whimsical charm and vibrant colors. These diverse styles utilize different textures and modeling techniques to evoke specific emotions and atmospheres. While currently prevalent in the movie and gaming industries, there is a growing need to democratize these applications, making them easily accessible and usable for consumers.

Prior works in 3D have focused primarily on altering textures, such as applying the color palette or patterns of one object onto another \cite{gatys2016image,kolkin2019style,chiu2020iterative,Ulyanov2016TextureNF,chaudhuri2021semi,Johnson2016PerceptualLF}. Another application domain is video, where the main challenge is maintaining temporal consistency across frames \cite{Xia2020RealtimeLP,Huang2017RealTimeNS}. While this can create visually interesting results, there is a need to explore how style transfer can be applied to both textures and geometry in 3D visuals.

Throughout art history, the study of style has encompassed various elements such as color, composition, and, notably, geometry and shape, which are considered crucial aspects in defining artistic identity \cite{Arnheim1967ArtAV,Gombrich1950TheSO}. In art, the technique of assemblage involves creating three-dimensional compositions on a defined substrate, akin to collage in two dimensions. Often utilizing found objects, assemblage offers a multidimensional approach to artistic expression. An illustrative example of this idea applied to oil paintings can be found in the works of Giuseppe Arcimboldo, who crafted human portraits from seasonal fruits, grains, and vegetables. Another example of this technique is the ``Mandolin'' sculpture by Pablo Picasso assembling the object from the wood pieces, see \cref{fig:idea_illustration}. Drawing inspiration from this approach, we propose a method where content object is assembled from various elements of the style scene.

In style transfer, the relationship between content and style representation is one of the central problems. The seminal work by Gatys \etal~\cite{gatys2016image} emphasized the optimization process that balances style and content loss to generate images, laying the foundation for subsequent investigations in the field. 
This paper laid the groundwork for subsequent explorations in the field, which have extended beyond two-dimensional texture transfer into three-dimensional shapes, facilitating the transfer of spatial structures between scenes.

The conventional optimization-based 2D style transfer matches style and content features distribution of the stylized image and style/content image respectively. Following this approach, we initially represent the content and style scenes as two-colored collections of particles using regularized Gaussian splatting (refer to \cref{sec:experiments} for details). To smoothly align these distributions, various methods can be used, but we choose to employ Sinkhorn divergence between the distributions of points corresponding to style and content scenes. Sinkhorn Divergence determins how the style scene should be adjusted to match the content scene. However, handling distributions with large sets of points becomes challenging. Therefore, we first segment the content scene into compact shapes and identify the best-fitting style segment. Then, we estimate the Sinkhorn divergence between pairs of style and content clusters, optimizing the selected style cluster to minimize the divergence between them. We opt for the Sinkhorn Divergence since it acts as a ``debiased'' Wasserstein-2 distance and is easier to tune.

In this study, we introduce a novel method for 3D scene style transfer, with the goal of accurately reproducing the geometry of the style scene within the content scene. Utilizing Sinkhorn divergence to adapt and integrate elements from the style scene, we construct a stylized content scene that preserves high visual fidelity and faithfully captures the geometric details of the style. Our qualitative experiments showcase that our approach outperforms existing methods, particularly in terms of geometric stylization quality.

\input{figures/fig_idea_illustration}

%% file: figures/fig_idea_illustration.tex
\begin{figure}[tb]
  \centering
  \begin{subfigure}{0.42\linewidth}
    \includegraphics[width=1.\textwidth]{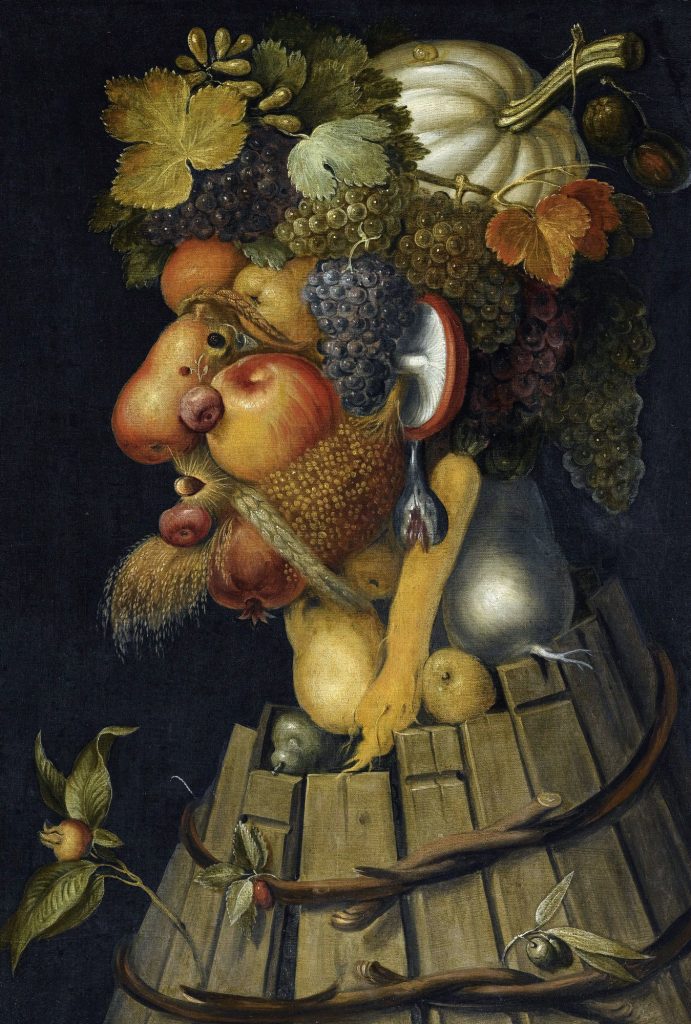}
    \caption{``Autumn'' by Guiseppe Arcimboldo.}
    \label{fig:arcimboldo_autumn}
  \end{subfigure}
  \hfill
  \begin{subfigure}{0.42\linewidth}
    \includegraphics[width=1.\textwidth]{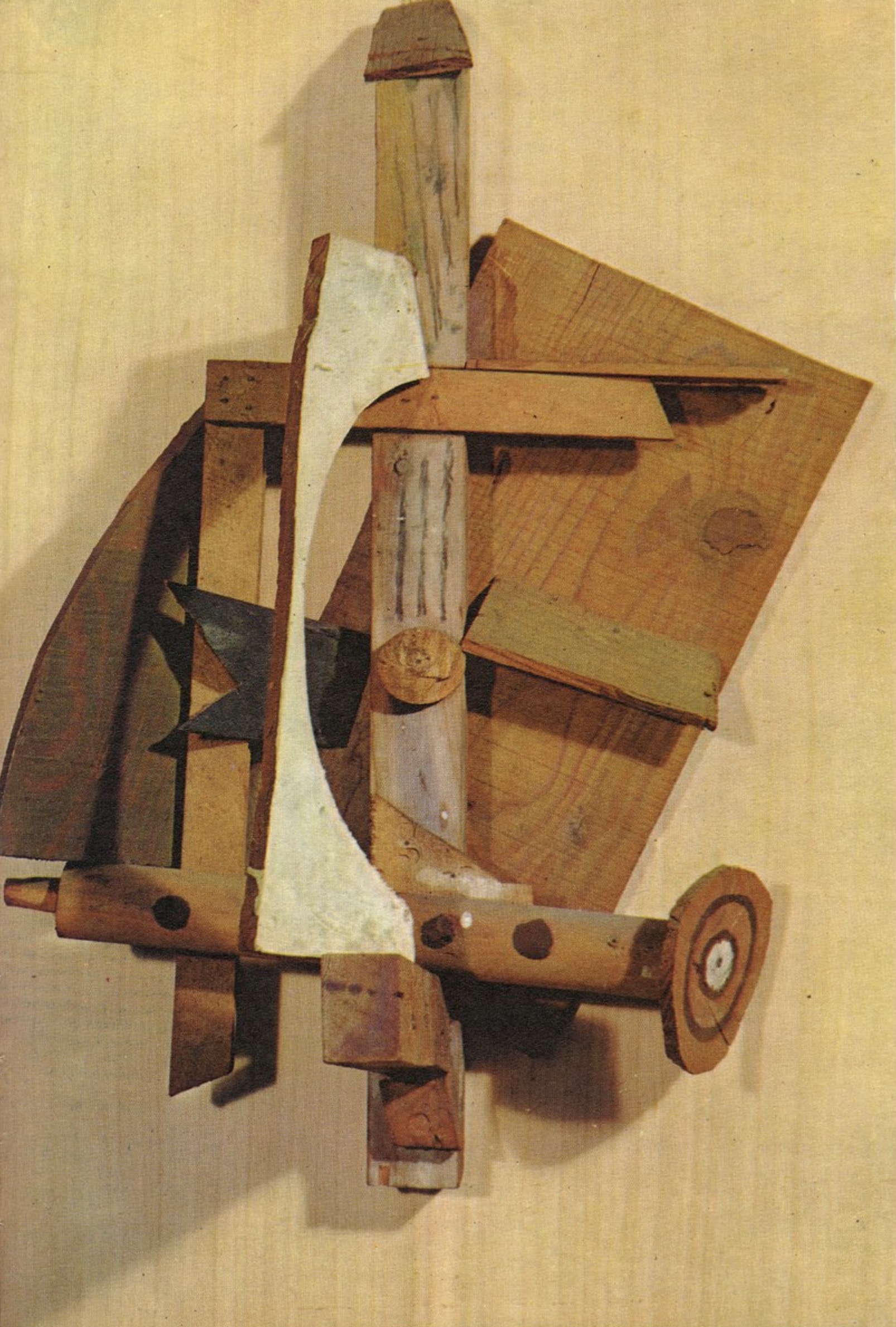}
    \caption{``Mandolin'' by Pablo Picasso.}
    \label{fig:arcimboldo_autumn}
  \end{subfigure}
  \caption{Two illustrations of the assemblage principle in art. The coarse level content can be constructed of local pieces of a style object of a different nature: vegetables and various pieces of wood. We follow a similar idea in our approach.}
  \label{fig:idea_illustration}
\end{figure}

%% file: tex/related_work.tex
\section{Related Work}

\subsubsection{Style Transfer}
Style transfer aims at generating images with the aesthetic style of the given style images while preserving their semantic content. Traditional methods use handcrafted features to simulate styles\cite{jacobs2001image}. Gatys \etal~\cite{gatys2016image} introduced neural style transfer in which the content and style representations are obtained as intermediate activations and the corresponding Gram matrices of a pre-trained VGG network~\cite{simonyan2014very}. The resulting image is generated by iteratively updating a random input until its content and style representations match the reference ones. Follow-up works \cite{Kotovenko2019ContentAS,li2016combining,liao2017visual,risser2017stable,gu2018arbitrary,kolkin2019style,mechrez2018contextual,Kotovenko2021RethinkingST} have explored alternative style loss formulations to improve the quality of semantic consistency and high-frequency style details. Rather than performing iterative optimization, feed-forward approaches \cite{huang2017arbitrary,Sanakoyeu2018ASC,Kotovenko2019ACT,an2021artflow,chiu2020iterative,park2019arbitrary,yao2019attention,sheng2018avatar,li2017universal} train neural networks that can capture the style information of the style image and transfer it to the input image with a single forward pass. These methods are faster than optimization-based techniques but often yield lower visual quality. Unlike neural style transfer methods that encode statistics of style features with a single Gram matrix, another line of work searches for nearest neighbors and minimizes distances between features extracted from corresponding content and style patches \cite{chen2016fast,li2016combining,liao2017visual,kolkin2022neural}. These methods achieve impressive 2D stylization quality when provided with source and target images that share similar semantics \cite{zhang2022arf}.

\subsection{3D style transfer}
In recent years many papers tried to transfer 2D results in 3D dimensions. The works can be generally split into two groups: with target style geometry and without one.

With the access to the style target geometry Nerf-Tex\cite{Baatz2022NeRFTexNR} fills the content scene with patches of the style example. A similar approach was adopted by \cite{TBSCB:2021:TFDM_tesselation-free}, where the style geometry is predefined using tessellation maps to articulate fine details in 3D geometry.  An alternative strategy involves leveraging depth maps alongside the original style image~\cite{Jung_2024_CVPR_GTfSRF}. However, these approaches encounter challenges when generalizing across diverse and intricate scenes with full 360-degree camera rotation, as the target depth is specified only for a single fixed viewpoint. 

Concurrently, a separate line of research focuses on modifying only the geometry of shapes without incorporating style textures~\cite{Segu20203DSNetUS, Wang2023CreativeBirds, kuznetsov2022_rendering_neural_materials_on_curved_surfaces}.

For methods lacking access to style geometry, the transformation is attempted based on 2D information using common 3D representations: point clouds~\cite{huang_2021_3d_scene_stylization_learning_to_stylize_novel_views}, meshes~\cite{li2024diffavatar, sarafianos2024garment3dgen}, NeRFs~\cite{wang2022nerf_nerf-art, liu2023stylerf, Xu_2023_CVPR, Chen2023TeSTNeRFT3, zhang2022arf, Nguyen_Phuoc_2022_SNeRF, refnerf} or  Gaussian Splats(GS)~\cite{gaussian_grouping}. Another approach to provide style information is by incorporating text prompt with desirable changes. Method~\cite{Chen2023TeSTNeRFT3} works using a text prompt indicating name of the artist in whose style the final image will be presented.
However, methods that take as an input image or text prompt barely change the geometry of the scenes. In majority of cases such approaches only mimic the colors of the style scene without altering the geometry.

%% file: tex/Approach.tex
\section{Methodology}\label{sec:methodology}

In this section, we describe the scene representation used throughout the paper, namely regularized Gaussian splattings. Following this, we introduce the Wasserstein-2 distance and its debiased counterpart, the Sinkhorn divergence, to quantify the divergence between two sets of Gaussian splattings. Finally, we outline WaSt-3D stylization algorithm, which is divided into three parts: a) cluster the content scene into disjoint subsets, b) select the best-fitting style cluster for each content cluster using the selection function $D$, and c) optimize style clusters to align with corresponding content clusters by minimizing the Sinkhorn divergence.

\subsection{Background: Gaussian splatting representation}
Neural Radiance Fields (NeRFs)~\cite{mildenhall2020nerf} have widely been utilized over the past few years as a neural network based representation that implicitly learns the scene density and view-dependent color. This representation allows for high-quality novel view synthesis once the NeRF model is optimized from a set of 2D input images. More recently, 3D Gaussian Splatting (3DGS)~\cite{kerbl3Dgaussians} has emerged as a powerful alternative that provides significant advantages in terms of optimization speed and real-time rendering while matching the state-of-the-art quality from NeRF-based approaches. 
3DGS achieves this by relying on explicit representation based on 3D Gaussians in contrast to the implicit representation of NeRF. To enable high-quality scene reconstruction, each Gaussian is characterized by a set of learnable parameters such as position $\mathbf{x} \in \mathbb{R}^3$, color $\mathbf{c} \in \mathbb{R}^3$, covariance matrix $\bm{\Sigma} \in \mathbb{SO}(3)$ (shape and orientation), scaling $\mathbf{S} \in \textbf{diag}(\mathbb{R}_+^3)$ and opacity $\alpha$. The rendering equation for a typical point-based approach is defined as:

\begin{equation}
	\label{eq:rendered_color}
	C = \sum_{i \in \mathcal{N}}
	\mathbf{c}_{i}\alpha_i
	\prod_{j=1}^{i-1}(1-\alpha_j).
\end{equation}
Here, $\mathcal{N}$ is the number of ordered points along the ray that are $\alpha$-blended per pixel and $\alpha_i$ is the blending opacity obtained by evaluating the 2D Gaussian projection at the pixel and multiplying it with the global opacity $\alpha$.

\subsection{Anisotropic Gaussian Splattings Representation}
\label{sec:anisotropic_reg}

Default 3D Gaussian splattings possess several parameters beyond coordinate and color, including scaling, rotation, and opacity. However, the objective of the training regime is often pixelwise similarity between the rendered image and the target image. This can inadvertently lead to individual Gaussians being stretched, resulting in undesirable needle-shaped artifacts protruding from the surface. This may result in unpleasant visual artifacts when we start splitting scenes into subsets with function $D_i$. To mitigate this effect, we introduce an anisotropic regularization for the individual Gaussian splattings, aiming to force them into spherical shapes. This is accomplished by minimizing the difference between the largest and smallest scaling components $g_\mathbf{S}$ of each Gaussian $g \in \mathcal{G}$:
\begin{equation}
    \mathcal{L}_{aniso} = \frac{1}{|\mathcal{G}|} \underset{g \in \mathcal{G}}{\sum} \left( \text{max}
     \left(\frac{\text{max} (g_\mathbf{S})}{\text{min} (g_\mathbf{S})}, r\right) - r \right),
    \label{eq:loss_pretraining_aniso}
\end{equation}
where the scalar $r$ determines how much can the largest and the smallest scaling parameter differ. 
Additionally we want to have Gaussians to have similar scale across the whole image.  

\begin{equation}
    \mathcal{L}_{unif} = \frac{1}{|\mathcal{G}|} \underset{g \in \mathcal{G}}{\sum} 
    \|g_\mathbf{S} - \hat{g_\mathbf{S}}\|^2,
    \label{eq:loss_pretraining_unif}
\end{equation}
where $\hat{g_\mathbf{S}}=(s,s,s) \in \mathbb{R}_+^3$ denotes target scale of Gaussians with $s\in\mathbb{R}_+$ being a positive real value indicating equal scale in every dimension.
Similar idea has been introduced as a regularization technique in \cite{xie2023physgaussian}. Moreover,  effect of the anisotropic regularization is discussed in the original paper on 3D Gaussian splattings \cite{kerbl3Dgaussians}.

\subsection{3D Style Transfer with Gaussian Splatting}

\subsubsection{Wasserstein-2 distance.}
The goal of the style transfer is to blend the style and content of two inputs across various domains such as images, audio, video, text, and 3D scenes. 
In this paper, we focus on blending two 3D scenes represented with 3D Gaussian splattings. We represent the style and content scenes by two collections of 3D Gaussian splattings, denoted as $\mathcal{G}_{\text{style}}$ and $\mathcal{G}_{\text{content}}$ respectively. The primary goal of stylization is to obtain a shape that locally resembles the style scene while globally resembling the content scene.

From a probabilistic standpoint, we need to ensure that the distributions of the style and content scenes are similar on a local scale. The distributions $p_c$ and $p_s$ of the Gaussians $\mathcal{G}_{\text{content}}$ and $\mathcal{G}_{\text{style}}$ lie on a compact connected Riemannian manifold $M$ with geodesic distance $d:M\times M \rightarrow \mathbb{R}_+$. We can use the Wasserstein-2 distance to find the optimal transport between them $\pi \in \Pi(p_s, p_c)$
    \begin{equation}
        \Pi(p_s, p_c) := \Bigl\{  \pi \in Prob(M\times M): \pi(\cdot,M)=p_c, \ \pi(M, \cdot)=p_s \Bigr\}.
        \label{eq:w2-distance}
    \end{equation}

    \begin{equation}
        \mathcal{W}_2 (p_s,p_c) := \Bigl[  \underset{\pi \in \Pi(p_s, p_c)}{\text{inf}}   
            \iint_{M\times M} d(x,y)^2 d\pi (x,y) \Bigr]^\frac{1}{2}.
        \label{eq:w2-distance}
    \end{equation}
This distance defines how dissimilar style and content distributions are. 

In practice, solving this problem is computationally intractable, necessitating regularization to make it applicable to the domain of Gaussian splatting in 3D spaces. One approach to encourage spread-out transportation plans $\pi$ is by adding the entropy term $H(\pi)$ to the Wasserstein-2 distance in \cref{eq:w2-distance}, resulting in an entropy-regularized objective function:

\begin{equation}
    \mathcal{W}_{2,\gamma}^2 (p_s,p_c) := \Bigl[  \underset{\pi \in \Pi(p_s, p_c)}{\text{inf}}   
        \iint_{M\times M} d(x,y)^2 d\pi (x,y) - \gamma H(\pi) \Bigr].
    \label{eq:w2-distance-entropy-reg}
\end{equation}

The entropy-regularized version of our objective function renders it strictly convex and is commonly referred to as the ``Schroedinger problem''. In practice, this regularization smoothens the problem and prevents overfitting. 
Varying the parameter $\gamma$ allows us to control the smoothness of the transport plan between the two distributions. A higher $\gamma$ value results in a smoother transport plan, where every Gaussian from the style scene is transported to every Gaussian from the content domain, approximating an ``average'' Gaussian. Conversely, a very small $\gamma$ value leads to each Gaussian being transported to just one Gaussian from the other domain, making the metric more specific.

We could have utilized~\cref{eq:w2-distance-entropy-reg} to compute the distance between two distributions and perform gradient flow to map one distribution to another. 
However, this is not a metric because $\mathcal{W}_{2,\gamma}^2 (p_s,p_s) \neq 0$ and $\mathcal{W}_{2,\gamma}^2 (p_c,p_c) \neq 0$. To address this, we employ the debiased version of the entropy regularized Wasserstein-2 distance dubbed Sinkhorn divergence:

\begin{equation}
    \mathcal{SD}_{2,\gamma}^2 (p_s,p_c) :=  \mathcal{W}_{2,\gamma}^2 (p_s,p_c) - 
     \mathcal{W}_{2,\gamma}^2 (p_s,p_s) -  \mathcal{W}_{2,\gamma}^2 (p_c,p_c).
    \label{eq:w2-sinkhorn}
\end{equation}
For values $\gamma \rightarrow \infty$ objective turns into $\frac{1}{2}MMD^2(p_s, p_c)$ , $\gamma \rightarrow 0$ objective turns into $\mathcal{W}_2^2 (p_s,p_c)$. See \cite{Ramdas2015OnWT,Peyr2018ComputationalOT}, where MMD stands for the Maximum Mean Discrepancy. 
This divergence between the distribution can be used to compute gradients and update one of the distributions. To estimate this value we use the Sinkhorn iterations method.

\subsubsection{Scene partitioning.} Estimating optimal transport between distributions containing hundreds of thousands or millions of points is not feasible, even with approximation algorithms. Thus, we tackle this problem by dividing it into smaller chunks and stylizing each part of the content scene separately. Although this may seem limiting, this approach guarantees faithful representation of the content scene by substituting it with small pieces of the style scene. Essentially, we want to ensure that our stylization locally resembles the style scene.

To achieve this, we divide the content scene $\mathcal{G}_{\text{content}}$ into a collection of clusters $C=\underset{i \in \{1,..,N\}}{\cup}C_i$, where each cluster consists of a collection of Gaussians $C_i=\{g | g \in \mathcal{G}_{\text{content}} \}$. Our objective is to transform these clusters while preserving the style and accurately representing the content scene. To achieve this, we first determine which style cluster fits best with each content cluster $C_i$.

This correspondence function $D_i$ between content clusters and parts of the style scene $\mathcal{G}_{\text{style}}$ is crucial for selecting the distributions $p_s$ and $p_c$ to be matched using the optimal transport plan.

To identify content clusters $C=\underset{i \in \{1,..,N\}}{\cup}C_i$, we perform K-Means clustering on the colored Gaussians. We then determine which style cluster should be replaced with the best-fitting style cluster by formulating a constrained optimization task. We fit each content cluster $C_i$ to the style scene, using only translation $\mathbf{t}_i \in \mathbb{R}^3$, rotation $\mathbf{R}_i \in \mathbb{SO}(3)$ ($\mathbb{SO}(3)$ is special orthogonal group of order $3$), and scaling $\mathbf{S}_i  \in \textbf{diag}(\mathbb{R}_+^3)$:

\begin{equation}
   \centering
  \begin{split}
    \text{min} & 
    \underset{g \in C_{i}}{\sum} \{ \text{min} 
    \Vert \mathbf{S}_i \mathbf{R}_i (g_{\textbf{x}} - \mathbf{t}_i) - g'_{\textbf{x}}\Vert_2\ |\  g' \in \mathcal{G}_{style} \} \\
     \text{s.t.} \ & \mathbf{R}_i \in \mathbb{SO}(3), 
                   \mathbf{S}_i  \in \textbf{diag}(\mathbb{R}_+^3), 
                   \mathbf{t}_i \in \mathbb{R}^3.
    \label{eq:stage1_opt}
  \end{split}    
\end{equation}
Thus, we minimize difference between coordinates $g_{\textbf{x}}$ of a cluster $C_i$ mapped using simple translation, rotation and scaling operation to the distribution of coordinates of the style scene $g'_{\textbf{x}}$. This constrained optimization problem can efficiently  partition the problem into a collection of smaller problems of minimizing  $\mathcal{SD}_{2,\gamma}^2 (C_i,C'_i)$ between every content cluster $C_i$ and assigned style cluster $C'_i$. 

Finally, we can define a selection mapping $D_i$ that given optimized parameters $\mathbf{t}_i, \mathbf{R}_i, \mathbf{S}_i$ selects Gaussians from the style scene $\mathcal{G}_{style}$.

\begin{equation}
    D_i: C_i \rightarrow \mathcal{G}_{style}, 
    D_i(C_i) = \underset{g \in C_i}{\cup}{\mathcal{N}_k(\mathbf{S}_i \mathbf{R}_i (g_{\textbf{x}} - \mathbf{t}_i))},
    \label{eq:cluster_selector}
\end{equation}
where $\mathcal{N}_k(g)$ denotes $k$ nearest neighbors of the Gaussian $g$ in the set of style Gaussians $\mathcal{G}_{style}$.

\subsubsection{Ultimate stylization objective.} With those preliminary steps we have successfully partitioned the problem into a collection of smaller OT problems that we can solve computing the Sinkhorn divergence between the distributions:

\begin{equation}
    \mathcal{L}_{opt} = \underset{i \in \{1,..,N\}}{\sum}{\mathcal{SD}_{2,\gamma}^2 (C_i,D_i(C_i))}.
    \label{eq:loss_final}
\end{equation}    

To elucidate the computation of the Sinkhorn Divergence on the set of Gaussians, we utilize both the coordinate and brightness attributes. This alignment ensures consistency not only in coordinates but also in the overall shading of the image between content clusters and their respective style clusters. In the experimental section we visualize different results obtained optimizing color or brightness components. The optimization process outlined in \cref{eq:stage1_opt} aims to minimize the loss over a set of parameters $\{\mathbf{t}_i, \mathbf{R}_i, \mathbf{S}_i \ |\ i \in {1,..,N}\}$, while \cref{eq:loss_final} optimizes over the color and coordinate components of the Gaussians $\{ D_i(C_i))\ |\ i \in {1,..,N}\}$.

%% file: tex/experiments.tex
\section{Experiments}\label{sec:experiments}

We conduct a comprehensive evaluation of WaSt-3D through a series of experiments. This includes qualitative and quantitative comparisons in~\cref{sec:qualquant} and further ablation studies in~\cref{sec:abl}. For high resolution video results please refer to the supplementary materials.

\subsubsection{Implementation Details.}
\label{sec:impl}

Our model is implemented in \texttt{PyTorch}. Initially, we pretrain content and style scenes using the standard Gaussian splatting training code, with additional regularization applied to the style scenes as specified in \cref{sec:anisotropic_reg}. All our experiments are conducted on a single A100 GPU, requiring $16$GB of VRAM for the largest style scenes containing $8e6$ Gaussians. On average our optimized scene contains $1e6-4e6$ Gaussians. The complete optimization pipeline takes approximately $8$ minutes given pretrained style and content scenes.

For Sinkhorn divergence computation, we utilize the GPU-optimized implementation available in the \texttt{geomloss} library, which relies on the \texttt{pykeops} library. Detailed algorithm is provided in the supplementary materials.

By default, we partition the content scene into $400$ clusters. In~\cref{fig:ablations} we show stylization results using fewer clusters. Prior to this partitioning, we sample points on the surface of the content scene to ensure that Gaussians are fitted only to the surface, mitigating issues stemming from the default Gaussian splatting fitting algorithm. This process is described in in the supplementary materials.

An additional step we need to perform is adjusting the scaling $\textbf{S}$ of individual Gaussians after they have been optimized with the loss defined in \cref{eq:loss_final}. We adjust their scale based on the changes in the average distance to the nearest neighbors. Further details are provided in the supplementary materials.

\subsubsection{Datasets.}

For our research, we need two type of 3D scenes: style and content. 
As content examples we used three publicly available datasets.
NeRF Synthetic~\cite{mildenhall2020nerf} is synthetic dataset of 3D objects with realistic non-Lambertian materials. For every scene it provides a set of 360 views and exact camera parameters. We also evaluated our dataset on real-world datasets with complex geometry LLFF~\cite{mildenhall2019llff} and Tanks\&Temples~\cite{Knapitsch2017TanksAndTemples}. Local Light Field Fusion(LLFF) is a bounded real world dataset captured with a handheld cellphone.  Tank\&Temples is a real world 360 scene dataset. The selected scenes exhibit diverse capture styles, including both bounded and unbounded settings. For style examples, we used scenes sourced from BlenderKit. All scenes are linked on the project page, and some are showcased in the supplementary material.

\input{figures/fig_comparison}

\subsubsection{Qualitative comparison.}
\label{sec:qualquant}

We compare WaSt-3D with three recent stylization approaches: ARF~\cite{zhang2022arf}, StyleRF~\cite{liu2023stylerf}
 and reimplemented by Ref-NPR~\cite{refnerf} SNeRF~\cite{Nguyen_Phuoc_2022_SNeRF}.
As illustrated in~\cref{fig:comparison}, our method excels in both high-quality stylization and preserving the essential features of content scenes. Demonstrating robustness,  WaSt-3D outperforms other methods across diverse content scenes and varying numbers of style scenes. Other methods fail to adequately preserve style details. For instance, when applied to a \textit{grass} style scene, only our method successfully maintains the intricate texture of the grass, showcasing the superior detail preservation achieved by our approach. More results are presented in~\cref{{fig:moreres}}, supplementary materials and on the project page.

\input{figures/fig_moreres}

\subsubsection{Quantitative comparison.}

\begin{table}[tb]
  \centering
  \caption{Quantitative Comparison. WaSt-3D surpasses other methods in both CLIP high-level details similarity and human preference score. Our model also takes less time to stylize a scene. Although it requires slightly more memory, this is a reasonable tradeo-ff considering the significant increase in stylization quality.}
  \begin{tabular}{ccccc}
    \toprule
    \multicolumn{1}{p{2cm}}{\ \textbf{Method}} &
    \multicolumn{1}{p{4cm}}{\centering \textbf{CLIP high-level} \\  \textbf{details similarity} $\uparrow$} & \multicolumn{1}{p{2cm}}{\centering \textbf{Human} \\  \textbf{preference} $\uparrow$} & 
    \multicolumn{1}{p{1.5cm}}{\ \textbf{Time} $\downarrow$ } &
    \multicolumn{1}{p{1.5cm}}{\textbf{VRAM} $\downarrow$ } \\
    \midrule
    ARF & \multicolumn{1}{c}{74.79 \%}  & \multicolumn{1}{c}{12.5\%} & 11 min & 9GB \\
    Style-RF & \multicolumn{1}{c}{74.94  \% } & \multicolumn{1}{c}{ 1.5 \%} & 18 min & \textbf{6GB} \\
    SNeRF & \multicolumn{1}{c}{76.99  \%} & \multicolumn{1}{c}{10.5 \%} & 30 min & 8GB \\
    \midrule
    Ours & \multicolumn{1}{c}{\textbf{84.40  \%}} & \multicolumn{1}{c}{\textbf{75.5 \%}}  & {\textbf{8 min}} & 16GB\\
    \bottomrule
  \end{tabular}
\label{tab:clip}
\end{table}

As demonstrated in~\cref{tab:clip},  WaSt-3D performance was evaluated using the CLIP~\cite{clip} similarity score. We extracted crops from the stylized scenes and compared them with crops from the original style scene to evaluate  ability of each method to preserve fine details. The results presented in Table~\ref{tab:clip} demonstrate that our method outperforms other methods in this regard.

Furthermore, to evaluate the stylization quality from a human perception standpoint, we conducted a user study. Participants were tasked with selecting the most appropriate and appealing stylization for pairs of content and style scenes. As shown in~\cref{tab:clip}, WaSt-3D achieved the highest scores in the user study. Additionally, we show in~\cref{tab:clip} speed and memory requirements of baselines in comparison to our method

\subsection{Ablation studies}
\label{sec:abl} 
\subsubsection{Style scene geometry preservation.}

To demonstrate the effectiveness of our approach in preserving the geometry of the original style scene, we compare the normals generated by our method with three alternative approaches: ARF\cite{zhang2022arf}, Style-RF\cite{liu2023stylerf} and SNeRF\cite{Nguyen_Phuoc_2022_SNeRF}. From~\cref{fig:normals} it is evident that our method faithfully reproduces the style geometry, while the other approaches introduce noise due to their optimization in RGB space. This highlights the robustness of our methodology in maintaining the integrity of the style scene geometry.
\input{figures/fig_normals}

\subsubsection{Sinkhorn divergence on different parameters.}
\label{sec:sink} 

In~\cref{sec:methodology}, we outlined the computation described in Equation~\ref{eq:loss_final}, which operates on various components of the Gaussians. Our standard configuration involves optimizing the coordinates $g_\textbf{x}$ and the luminance derived from $g_\textbf{c}$ using the formula $(0.299 \times R + 0.587 \times G + 0.114 \times B)$, where $R$, $G$, and $B$ denote the values of the red, green, and blue channels within the range $[0;1]$. As an alternative, we conducted experiments solely on the positional values $g_\textbf{x}$ and on both the positional and color values $g_\textbf{c}$. The outcomes of these experiments are depicted in \cref{fig:color_vs_brightness_vs_coords}. It is noteworthy that optimizing brightness (the second column) aids in preserving the original scene shading, thereby enhancing the portrayal of the scene volume.

We visualize effect of the entropy regularization term $\gamma$ in the supplementary material, on default we use $\gamma=0.05$. 

\input{figures/fig_color_vs_brightness_vs_coords}

\subsubsection{Wasserstein distance alternatives.}
\label{sec:wd_al}

To prevent overstretching or tearing of the style cluster we can use the surface energy regularization. This mapping $D$ allows translation, rotation, scaling, and stretching of individual Gaussians independently of each other. However, not controlling this operation we can mess up the original style cluster. To prevent too strong stretching of the style cluster $C$ we compute the surface energy of every cluster before and after the transformation:

\begin{equation}
    E(C, D(C)):=\underset{g \in C}{\sum} \underset{h \in \mathcal{N}(g)}{\sum} \left\Vert \Vert g-h \Vert_2 - \frac{1}{\lambda_D^2} \Vert D(g) - D(h) \Vert_2\right\Vert_1. 
    \label{eq:surface_energy}
\end{equation}

This loss is inspired by the ARAP loss~\cite{ARAP_modeling:2007} that is used to model the elastic deformation of meshes, $\lambda_D$ is the scaling constant of operator $D$. By constructing the function and surface energy  we guarantee that  the style will not be corrupted under any of the following transformation: translation, rotation, scaling and stretching while preserving surface energy.  The results of this loss are depicted in the right column of \cref{fig:ablations}.
\input{figures/fig_ablations}: content becomes less visible and follows the content shape poorly.
We additionally replace our loss defined in \cref{eq:loss_final} with the entropy regularized Wasserstein-2 distance \cref{eq:w2-distance-entropy-reg}, this makes the optimization less stable in practice and requires more hyperparameters tweaking. Results are provided in the supplementary.

%% file: figures/fig_comparison.tex
\begin{figure}
    \centering
    \includegraphics[width=1.0\textwidth]{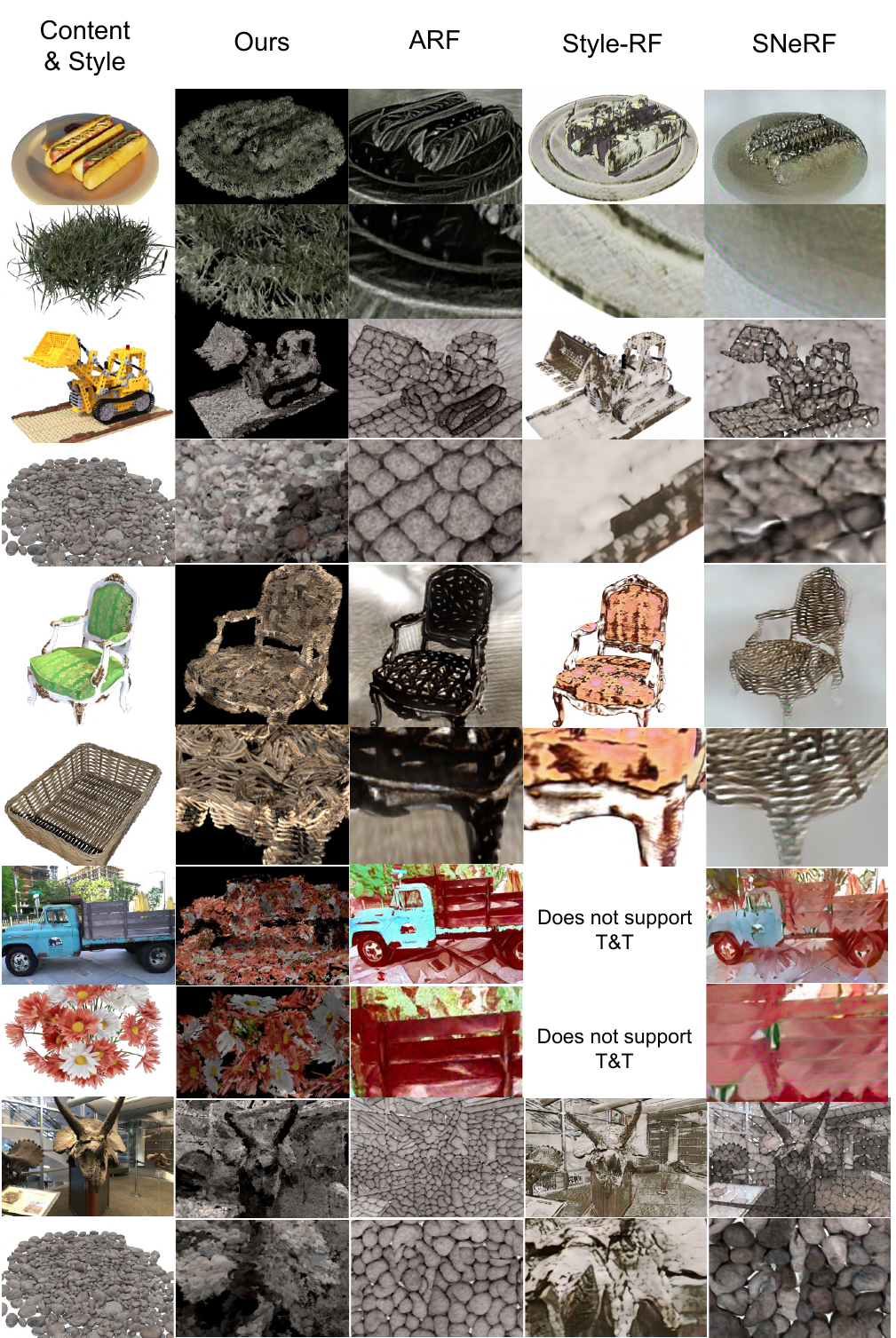}
    \caption{Comparison of WaSt-3D against three different approaches: ARF\cite{zhang2022arf}, SNeRF\cite{Nguyen_Phuoc_2022_SNeRF}, and StyleRF\cite{liu2023stylerf}. 
    We conducted the comparisons on three NeRF Synthetic scenes: \textit{hotdog}, \textit{lego}, and \textit{chair}; one scene, \textit{truck}, from the Tanks\&Temples dataset\cite{Knapitsch2017TanksAndTemples}; and the scene \textit{horns} from the LLFF dataset\cite{mildenhall2019llff}. Style scenes are named \textit{grass}, \textit{pebbles}, \textit{wicker basket}, \textit{bouquet}, and \textit{stones}; examples of these style scenes are provided in the supplementary materials. For each image, we provide a close-up of the same fragment across different methods. It is important to note that while overall stylization seems comparable for small images, our method delivers favorable results for high-resolution images. Moreover, our method preserves the 3D geometry of the style scene. Please refer to the project page for video renderings of the scenes in high resolution.}
    \label{fig:comparison}
\end{figure}

%% file: figures/fig_moreres.tex
\begin{figure}
    \centering
    \includegraphics[width=1\textwidth]{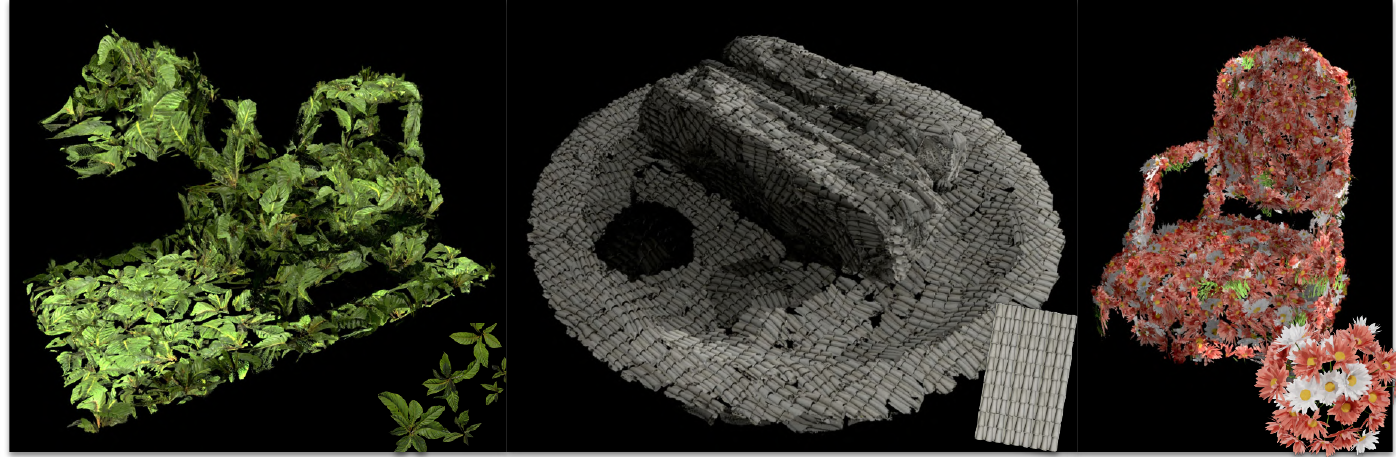}
    \caption{Additional visual results for our method on NeRF Synthetic objects~\cite{mildenhall2020nerf}. The style scene for each object is depicted in the lower right angle for optimal presentation. }
    \label{fig:moreres}
\end{figure}

%% file: figures/fig_normals.tex
\begin{figure}[t]
    \centering
    \includegraphics[width=1\textwidth]{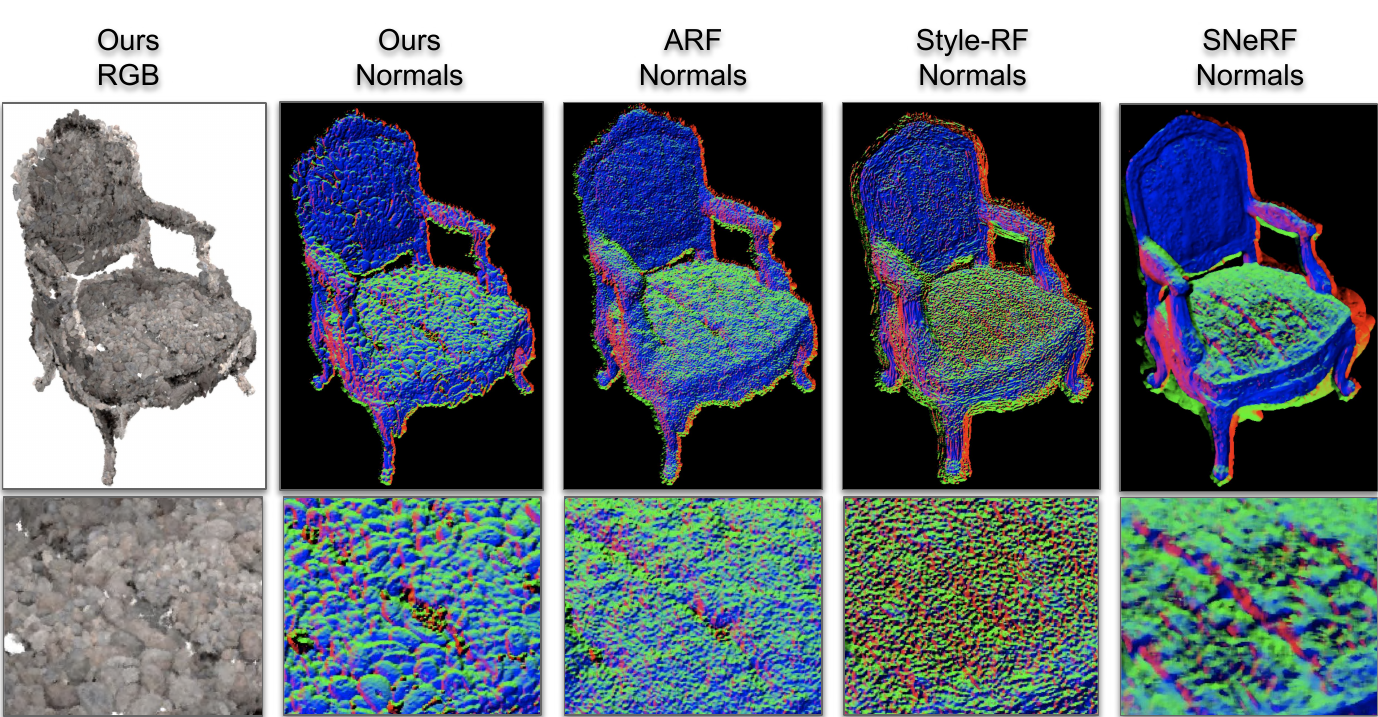}
    \caption{Visualization of the normals obtained after stylization of the \textit{chair} scene using the \textit{pebbles} style with different methods: ARF\cite{zhang2022arf}, Style-RF\cite{liu2023stylerf} and SNeRF\cite{Nguyen_Phuoc_2022_SNeRF}.}
    \label{fig:normals}
\end{figure}

%% file: figures/fig_color_vs_brightness_vs_coords.tex
\begin{figure}
    \centering
    \includegraphics[width=1\textwidth]{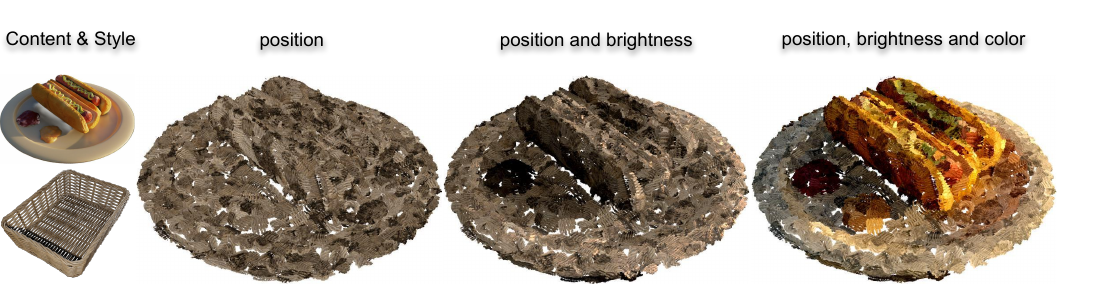}
    \caption{Results of the stylization when minimizing $\mathcal{L}_{opt}$:  for positions only,  for position and brightness, for position, brightness and color. Original content scene \textit{hotdog} and  style scene \textit{wicker basket} is provided in the first column.}
    \label{fig:color_vs_brightness_vs_coords}
\end{figure}

%% file: figures/fig_ablations.tex
\begin{figure}
    \centering 
    \includegraphics[width=1\textwidth]{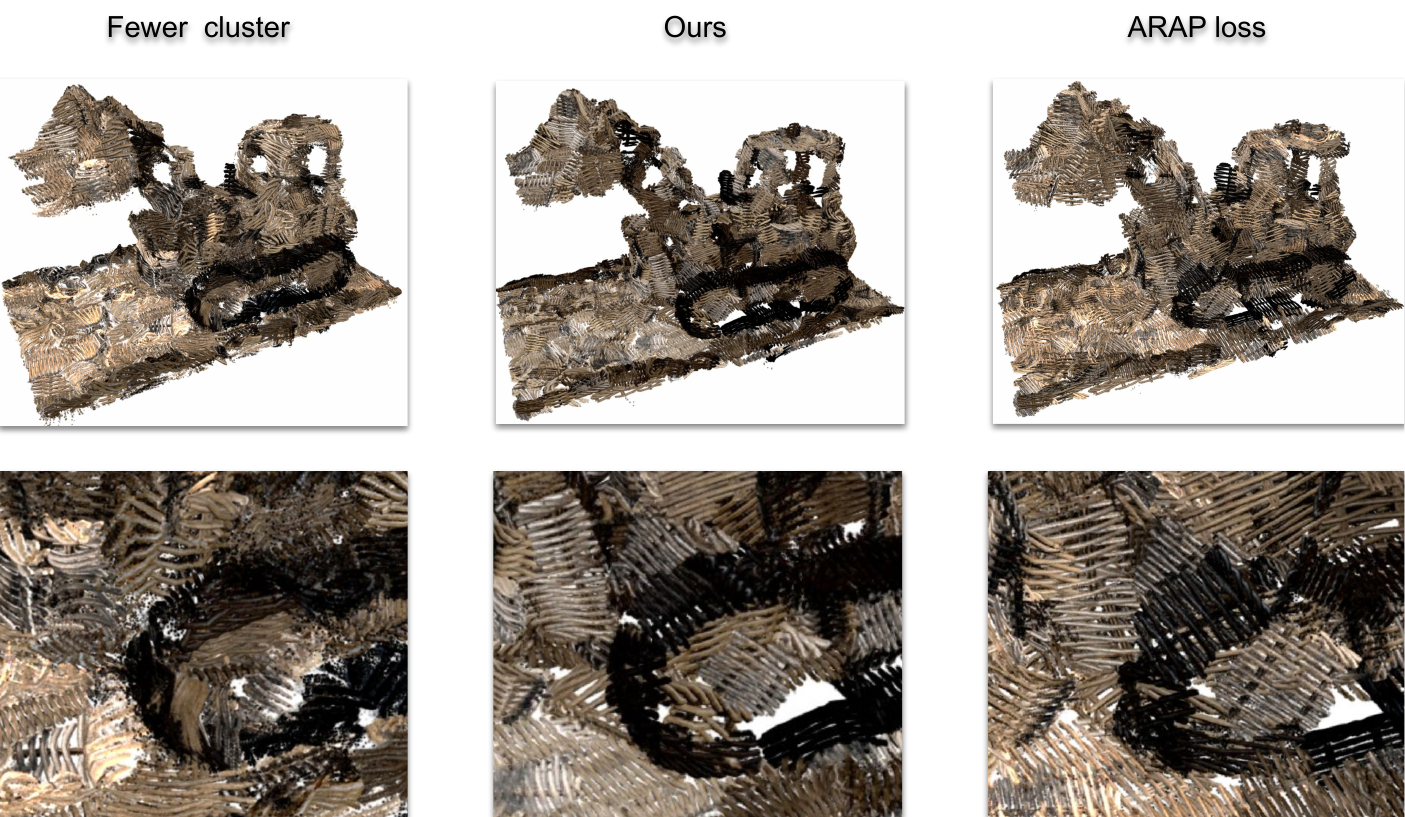}
    \caption{We ablate two crucial components of WaSt-3D. In the middle we show our main model stylizing \textit{lego} scene using \textit{wicker basket} scene. On the right side we replace Sinkhorn divergence with the ARAP~\cite{ARAP_modeling:2007} loss. On the left side we reduce number of content cluster $N$ from $400$ to $200$.}
    \label{fig:ablations}
\end{figure}

%% file: tex/conclusion.tex
\section{Conclusion}
In this paper, we propose the WaSt-3D method for stylizing 3D scenes using another 3D style scene as a reference. Recognizing the difficulty of general scene-to-scene translation, we first partition the content scene into simpler components, identify the best-fitting style part for each content cluster, and fine-tune the style by optimizing the Wasserstein-2 distance for each pair of clusters. This approach to matching distributions, coupled with a robust representation of the style scenes, enables a faithful reproduction of the style scene geometry. This stands in contrast to the classical approach of rendering scenes to align with feature distributions in the image encoder space, distinguishing our method from conventional 3D scene processing techniques. We conduct ablation studies on various components of our model and assess the impact of different parameters on its final performance. Additionally, we demonstrate the effectiveness of our approach through user preference studies and by measuring the CLIP similarity score of the optimized scene and the style scene patches. We believe that our approach holds promise for the community and can pave the way for alternative research directions in this field.

%% file: wast3d.bbl
\begin{thebibliography}{10}
\providecommand{\url}[1]{\texttt{#1}}
\providecommand{\urlprefix}{URL }
\providecommand{\doi}[1]{https://doi.org/#1}

\bibitem{an2021artflow}
An, J., Huang, S., Song, Y., Dou, D., Liu, W., Luo, J.: Artflow: Unbiased image style transfer via reversible neural flows. In: Proceedings of the IEEE/CVF Conference on Computer Vision and Pattern Recognition. pp. 862--871 (2021)

\bibitem{Arnheim1967ArtAV}
Arnheim, R.: Art and visual perception, a psychology of the creative eye (1967)

\bibitem{Baatz2022NeRFTexNR}
Baatz, H., Granskog, J., Papas, M., Rousselle, F., Nov{\'a}k, J.: Nerf‐tex: Neural reflectance field textures. Computer Graphics Forum  \textbf{41} (2022)

\bibitem{chaudhuri2021semi}
Chaudhuri, B., Sarafianos, N., Shapiro, L., Tung, T.: Semi-supervised synthesis of high-resolution editable textures for 3d humans. In: CVPR (2021)

\bibitem{Chen2023TeSTNeRFT3}
Chen, J., Ji, B., Zhang, Z., Chu, T., Zuo, Z., Zhao, L., Xing, W., Lu, D.: Testnerf: Text-driven 3d style transfer via cross-modal learning. In: International Joint Conference on Artificial Intelligence (2023)

\bibitem{chen2016fast}
Chen, T.Q., Schmidt, M.: Fast patch-based style transfer of arbitrary style. arXiv preprint arXiv:1612.04337  (2016)

\bibitem{chiu2020iterative}
Chiu, T.Y., Gurari, D.: Iterative feature transformation for fast and versatile universal style transfer. In: Computer Vision--ECCV 2020: 16th European Conference, Glasgow, UK, August 23--28, 2020, Proceedings, Part XIX 16. pp. 169--184. Springer (2020)

\bibitem{gatys2016image}
Gatys, L.A., Ecker, A.S., Bethge, M.: Image style transfer using convolutional neural networks. In: Proceedings of the IEEE conference on computer vision and pattern recognition. pp. 2414--2423 (2016)

\bibitem{Gombrich1950TheSO}
Gombrich, E.H.: The story of art (1950)

\bibitem{gu2018arbitrary}
Gu, S., Chen, C., Liao, J., Yuan, L.: Arbitrary style transfer with deep feature reshuffle. In: Proceedings of the IEEE Conference on Computer Vision and Pattern Recognition. pp. 8222--8231 (2018)

\bibitem{Huang2017RealTimeNS}
Huang, H., Wang, H., Luo, W., Ma, L., Jiang, W., Zhu, X., Li, Z., Liu, W.: Real-time neural style transfer for videos. 2017 IEEE Conference on Computer Vision and Pattern Recognition (CVPR) pp. 7044--7052 (2017)

\bibitem{huang_2021_3d_scene_stylization_learning_to_stylize_novel_views}
Huang, H.P., Tseng, H.Y., Saini, S., Singh, M., Yang, M.H.: Learning to stylize novel views. In: ICCV (2021)

\bibitem{huang2017arbitrary}
Huang, X., Belongie, S.: Arbitrary style transfer in real-time with adaptive instance normalization. In: Proceedings of the IEEE international conference on computer vision. pp. 1501--1510 (2017)

\bibitem{jacobs2001image}
Jacobs, C., Salesin, D., Oliver, N., Hertzmann, A., Curless, A.: Image analogies. In: Proceedings of Siggraph. pp. 327--340 (2001)

\bibitem{Johnson2016PerceptualLF}
Johnson, J., Alahi, A., Fei-Fei, L.: Perceptual losses for real-time style transfer and super-resolution. ArXiv  \textbf{abs/1603.08155} (2016)

\bibitem{Jung_2024_CVPR_GTfSRF}
Jung, H., Nam, S., Sarafianos, N., Yoo, S., Sorkine-Hornung, A., Ranjan, R.: Geometry transfer for stylizing radiance fields. In: Proceedings of the IEEE/CVF Conference on Computer Vision and Pattern Recognition (CVPR). pp. 8565--8575 (June 2024)

\bibitem{kerbl3Dgaussians}
Kerbl, B., Kopanas, G., Leimk{\"u}hler, T., Drettakis, G.: 3d gaussian splatting for real-time radiance field rendering. ACM Transactions on Graphics  \textbf{42}(4) (July 2023)

\bibitem{Knapitsch2017TanksAndTemples}
Knapitsch, A., Park, J., Zhou, Q.Y., Koltun, V.: Tanks and temples: Benchmarking large-scale scene reconstruction. ACM Transactions on Graphics  \textbf{36}(4) (2017)

\bibitem{kolkin2022neural}
Kolkin, N., Kucera, M., Paris, S., Sykora, D., Shechtman, E., Shakhnarovich, G.: Neural neighbor style transfer. arXiv e-prints pp. arXiv--2203 (2022)

\bibitem{kolkin2019style}
Kolkin, N., Salavon, J., Shakhnarovich, G.: Style transfer by relaxed optimal transport and self-similarity. In: Proceedings of the IEEE/CVF Conference on Computer Vision and Pattern Recognition. pp. 10051--10060 (2019)

\bibitem{Kotovenko2019ContentAS}
Kotovenko, D., Sanakoyeu, A., Lang, S., Ommer, B.: Content and style disentanglement for artistic style transfer. 2019 IEEE/CVF International Conference on Computer Vision (ICCV) pp. 4421--4430 (2019)

\bibitem{Kotovenko2019ACT}
Kotovenko, D., Sanakoyeu, A., Ma, P., Lang, S., Ommer, B.: A content transformation block for image style transfer. 2019 IEEE/CVF Conference on Computer Vision and Pattern Recognition (CVPR) pp. 10024--10033 (2019)

\bibitem{Kotovenko2021RethinkingST}
Kotovenko, D., Wright, M., Heimbrecht, A., Ommer, B.: Rethinking style transfer: From pixels to parameterized brushstrokes. 2021 IEEE/CVF Conference on Computer Vision and Pattern Recognition (CVPR) pp. 12191--12200 (2021)

\bibitem{kuznetsov2022_rendering_neural_materials_on_curved_surfaces}
Kuznetsov, A., Wang, X., Mullia, K., Luan, F., Xu, Z., Ha\v{s}an, M., Ramamoorthi, R.: Rendering neural materials on curved surfaces. SIGGRAPH '22 Conference Proceedings  (2022)

\bibitem{li2016combining}
Li, C., Wand, M.: Combining markov random fields and convolutional neural networks for image synthesis. In: Proceedings of the IEEE conference on computer vision and pattern recognition. pp. 2479--2486 (2016)

\bibitem{li2024diffavatar}
Li, Y., Chen, H.y., Larionov, E., Sarafianos, N., Matusik, W., Stuyck, T.: Diffavatar: Simulation-ready garment optimization with differentiable simulation. In: CVPR. pp. 4368--4378 (2024)

\bibitem{li2017universal}
Li, Y., Fang, C., Yang, J., Wang, Z., Lu, X., Yang, M.H.: Universal style transfer via feature transforms. Advances in neural information processing systems  \textbf{30} (2017)

\bibitem{liao2017visual}
Liao, J., Yao, Y., Yuan, L., Hua, G., Kang, S.B.: Visual attribute transfer through deep image analogy. arXiv preprint arXiv:1705.01088  (2017)

\bibitem{liu2023stylerf}
Liu, K., Zhan, F., Chen, Y., Zhang, J., Yu, Y., Saddik, A.E., Lu, S., Xing, E.: Stylerf: Zero-shot 3d style transfer of neural radiance fields. In: CVPR (2023)

\bibitem{mechrez2018contextual}
Mechrez, R., Talmi, I., Zelnik-Manor, L.: The contextual loss for image transformation with non-aligned data. In: Proceedings of the European conference on computer vision (ECCV). pp. 768--783 (2018)

\bibitem{mildenhall2019llff}
Mildenhall, B., Srinivasan, P.P., Ortiz-Cayon, R., Kalantari, N.K., Ramamoorthi, R., Ng, R., Kar, A.: Local light field fusion: Practical view synthesis with prescriptive sampling guidelines. ACM Transactions on Graphics (TOG)  (2019)

\bibitem{mildenhall2020nerf}
Mildenhall, B., Srinivasan, P.P., Tancik, M., Barron, J.T., Ramamoorthi, R., Ng, R.: Nerf: Representing scenes as neural radiance fields for view synthesis. In: ECCV (2020)

\bibitem{Nguyen_Phuoc_2022_SNeRF}
Nguyen-Phuoc, T., Liu, F., Xiao, L.: Snerf: stylized neural implicit representations for 3d scenes. ACM Transactions on Graphics  \textbf{41}(4),  1–11 (Jul 2022)

\bibitem{park2019arbitrary}
Park, D.Y., Lee, K.H.: Arbitrary style transfer with style-attentional networks. In: proceedings of the IEEE/CVF conference on computer vision and pattern recognition. pp. 5880--5888 (2019)

\bibitem{Peyr2018ComputationalOT}
Peyr{\'e}, G., Cuturi, M.: Computational optimal transport. Found. Trends Mach. Learn.  \textbf{11},  355--607 (2018)

\bibitem{clip}
Radford, A., Kim, J.W., Hallacy, C., Ramesh, A., Goh, G., Agarwal, S., Sastry, G., Askell, A., Mishkin, P., Clark, J., Krueger, G., Sutskever, I.: Learning transferable visual models from natural language supervision (2021)

\bibitem{Ramdas2015OnWT}
Ramdas, A., Trillos, N.G., Cuturi, M.: On wasserstein two-sample testing and related families of nonparametric tests. Entropy  \textbf{19}, ~47 (2015)

\bibitem{risser2017stable}
Risser, E., Wilmot, P., Barnes, C.: Stable and controllable neural texture synthesis and style transfer using histogram losses. arXiv preprint arXiv:1701.08893  (2017)

\bibitem{Sanakoyeu2018ASC}
Sanakoyeu, A., Kotovenko, D., Lang, S., Ommer, B.: A style-aware content loss for real-time hd style transfer. ArXiv  \textbf{abs/1807.10201} (2018)

\bibitem{sarafianos2024garment3dgen}
Sarafianos, N., Stuyck, T., Xiang, X., Li, Y., Popovic, J., Ranjan, R.: {Garment3DGen}: {3D} garment stylization and texture generation. arXiv preprint arXiv:2403.18816  (2024)

\bibitem{Segu20203DSNetUS}
Segu, M., Grinvald, M., Siegwart, R.Y., Tombari, F.: 3dsnet: Unsupervised shape-to-shape 3d style transfer. ArXiv  \textbf{abs/2011.13388} (2020)

\bibitem{sheng2018avatar}
Sheng, L., Lin, Z., Shao, J., Wang, X.: Avatar-net: Multi-scale zero-shot style transfer by feature decoration. In: Proceedings of the IEEE conference on computer vision and pattern recognition. pp. 8242--8250 (2018)

\bibitem{simonyan2014very}
Simonyan, K., Zisserman, A.: Very deep convolutional networks for large-scale image recognition. arXiv preprint arXiv:1409.1556  (2014)

\bibitem{ARAP_modeling:2007}
Sorkine, O., Alexa, M.: As-rigid-as-possible surface modeling. In: Proceedings of EUROGRAPHICS/ACM SIGGRAPH Symposium on Geometry Processing. pp. 109--116 (2007)

\bibitem{TBSCB:2021:TFDM_tesselation-free}
Thonat, T., Beaune, F., Sun, X., Carr, N., Boubekeur, T.: Tessellation-free displacement mapping for ray tracing  \textbf{40}(6) (dec 2021)

\bibitem{Ulyanov2016TextureNF}
Ulyanov, D., Lebedev, V., Vedaldi, A., Lempitsky, V.S.: Texture networks: Feed-forward synthesis of textures and stylized images. ArXiv  \textbf{abs/1603.03417} (2016)

\bibitem{wang2022nerf_nerf-art}
Wang, C., Jiang, R., Chai, M., He, M., Chen, D., Liao, J.: Nerf-art: Text-driven neural radiance fields stylization. arXiv preprint arXiv:2212.08070  (2022)

\bibitem{Wang2023CreativeBirds}
Wang, R., Que, G., Chen, S., Li, X., Li, J.Y., Yang, J.: Creative birds: Self-supervised single-view 3d style transfer. ArXiv  \textbf{abs/2307.14127} (2023)

\bibitem{Xia2020RealtimeLP}
Xia, X., Xue, T., Lai, W.S., Sun, Z., Chang, A., Kulis, B., Chen, J.: Real-time localized photorealistic video style transfer. 2021 IEEE Winter Conference on Applications of Computer Vision (WACV) pp. 1088--1097 (2020)

\bibitem{xie2023physgaussian}
Xie, T., Zong, Z., Qiu, Y., Li, X., Feng, Y., Yang, Y., Jiang, C.: Physgaussian: Physics-integrated 3d gaussians for generative dynamics. arXiv preprint arXiv:2311.12198  (2023)

\bibitem{Xu_2023_CVPR}
Xu, S., Li, L., Shen, L., Lian, Z.: Desrf: Deformable stylized radiance field. In: Proceedings of the IEEE/CVF Conference on Computer Vision and Pattern Recognition (CVPR) Workshops. pp. 709--718 (June 2023)

\bibitem{yao2019attention}
Yao, Y., Ren, J., Xie, X., Liu, W., Liu, Y.J., Wang, J.: Attention-aware multi-stroke style transfer. In: Proceedings of the IEEE/CVF conference on computer vision and pattern recognition. pp. 1467--1475 (2019)

\bibitem{gaussian_grouping}
Ye, M., Danelljan, M., Yu, F., Ke, L.: Gaussian grouping: Segment and edit anything in 3d scenes. arXiv preprint arXiv:2312.00732  (2023)

\bibitem{zhang2022arf}
Zhang, K., Kolkin, N., Bi, S., Luan, F., Xu, Z., Shechtman, E., Snavely, N.: Arf: Artistic radiance fields. In: European Conference on Computer Vision. pp. 717--733. Springer (2022)

\bibitem{refnerf}
Zhang, Y., He, Z., Xing, J., Yao, X., Jia, J.: Ref-npr: Reference-based non-photorealistic radiance fields for controllable scene stylization. In: Proceedings of the IEEE/CVF Conference on Computer Vision and Pattern Recognition (CVPR). pp. 4242--4251 (June 2023)

\end{thebibliography}
